\documentclass[letterpaper, 10 pt, conference]{ieeeconf}  

\IEEEoverridecommandlockouts                              

\usepackage{cite}
\usepackage{multicol}
\usepackage{amsmath}
\usepackage{booktabs}
\usepackage{tabularx}
\usepackage{subcaption}
\usepackage{graphicx}
\usepackage{censor}
\usepackage[table]{xcolor}
\usepackage[bookmarks=true]{hyperref}

\title{\LARGE \bf
Micro-Localization: High-Precision Mobile Robot Localization Using Microscopic Ground Texture
}

\author{Aaron Wilhelm$^{1}$, Marina de Albuquerque Marcon$^{2}$,
Emith Uyanwatte$^{1}$, and Nils Napp$^{1}$%
\thanks{$^{1}$School of Electrical and Computer Engineering,
Cornell University, Ithaca, NY 14853, USA.
Corresponding author: Aaron Wilhelm
(\texttt{ajw344@cornell.edu}).}%
\thanks{$^{2}$Department of Mechanical Engineering,
University of Michigan, Ann Arbor, MI 48109, USA.}%
\thanks{This work has been submitted to the IEEE for possible publication.
Copyright may be transferred without notice, after which this version
may no longer be accessible.}%
}

\begin{document}
\bstctlcite{IEEEexample:BSTcontrol}

\maketitle
\thispagestyle{empty}
\pagestyle{empty}

\begin{abstract}
We demonstrate high-precision global localization using natural ground texture viewed through microscope optics, achieving localization in the tens-of-micrometers regime.
For the planar ground texture localization problem considered here, we relate optical magnification and image-space localization uncertainty to physical localization precision.
Leveraging this relationship, we then develop a microscope-based localization system that addresses the calibration, camera-modeling, and focus challenges introduced by high-magnification imaging.
We evaluate this system across several surface textures using a calibrated precision gantry.
Across 1,007 held-out queries per surface, the system localizes every query and achieves translation RMSEs of 28--73~$\mu$m.
Independent photometric registration agrees with the feature-based localization estimates to within approximately 9~$\mu$m RMSE.
We further demonstrate localization on a low-cost mobile robot using the gantry-collected maps, and also show that the system can resolve micrometer-scale 
displacement caused by thermal expansion.
These results show that microscopic natural surface texture can bring mobile localization into a precision regime typically associated with fixed positioning systems.
\end{abstract}

\section{Introduction}

Mobile robots offer a compelling way to bring inspection, manipulation, and fabrication capabilities to large or reconfigurable workspaces.
However, many of these tasks require positioning accuracy far finer than that needed for navigation.
Precision manufacturing systems such as CNC machines can achieve micrometer-scale positioning, but their rigid mechanical structure confines them to a fixed work envelope.
In many mobile applications, such precision may only be needed near the task, motivating localization methods that provide high accuracy locally while preserving the flexibility of a mobile platform.

Ground texture localization provides a promising basis for this capability.
Rather than relying on sparse landmarks or installed fiducials, these methods use the natural surface texture beneath a robot as a global reference.
Because the surface is viewed at close range, a given planar displacement produces a larger displacement of the texture in the native sensor image than it would under a more distant viewing geometry, helping enable high physical localization precision.
We refer to this relationship between physical displacement and the resulting image displacement as the geometric sensitivity of the imaging system.
Building on this principle, we use microscope optics to substantially increase optical magnification to target globally referenced localization in the tens-of-micrometers regime.

\begin{figure}
    \centering
    \includegraphics[width=0.98\linewidth]{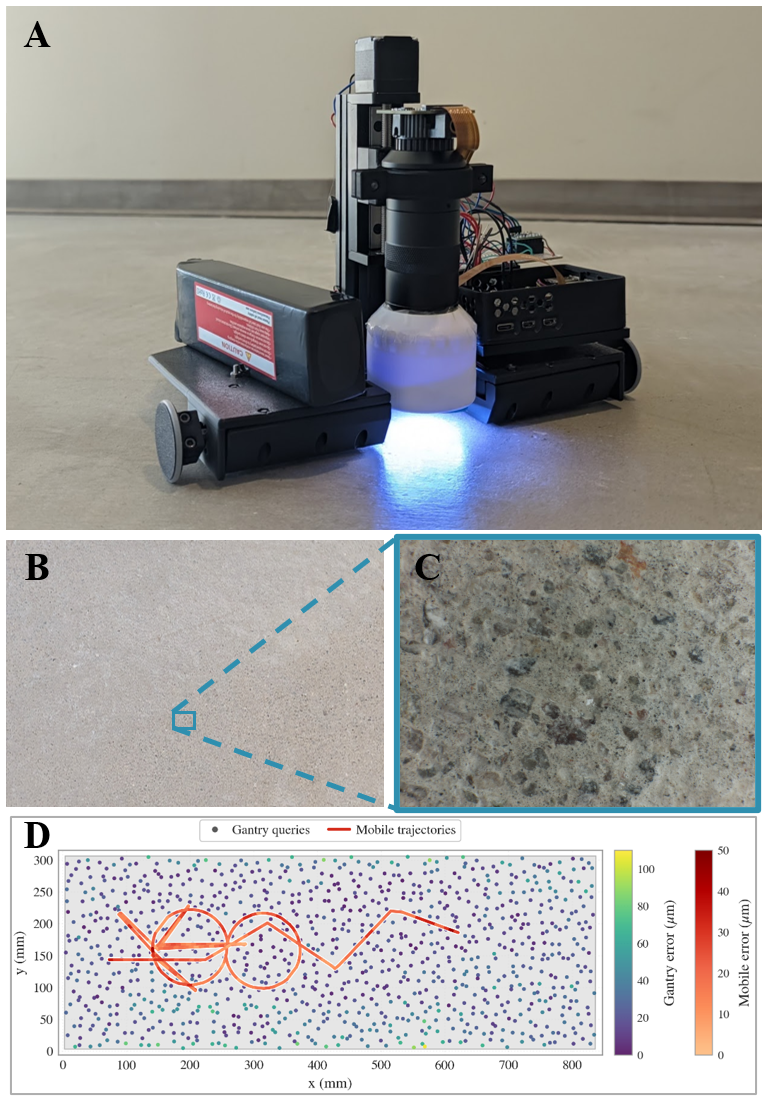}
    \caption{\textbf{System Hardware, Multi-Scale Texture, and Localization Performance}. 
    (A) Custom mobile platform equipped with the downward-facing camera, microscope optics, and illumination.
    (B) Macroscopic view of concrete and (C) the corresponding microscopic view, revealing distinctive surface structure used for localization.
    (D) Spatial localization performance over the mapped workspace, showing gantry query errors and mobile-robot trajectories colored by translational error.}
    \label{fig:robot_figure}
    \vspace{-8pt}
\end{figure}

\begin{figure}[tbhp]
    \centering
    \vspace{5pt}
    \includegraphics[width=0.98\columnwidth]{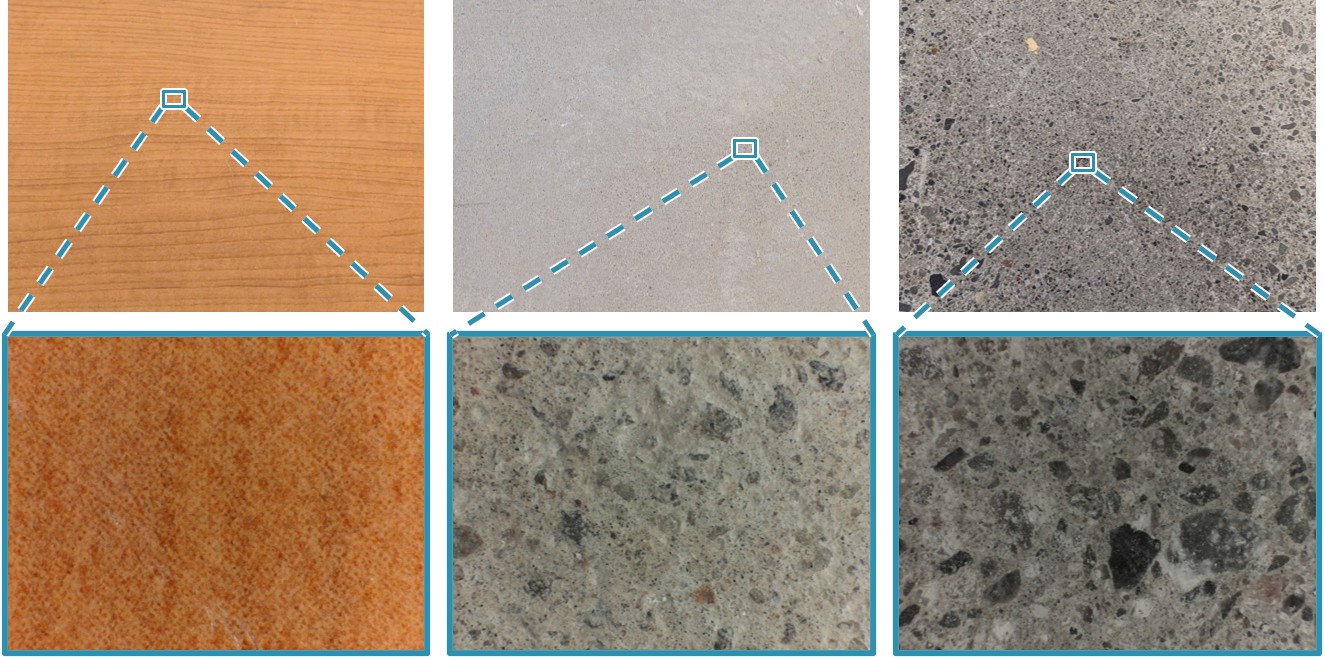}
    \caption{\textbf{Macroscopic versus Microscopic Texture Characteristics}. 
    Macroscopic views (top) and corresponding microscopic views (bottom) of printed wood grain (left), concrete (middle), and terrazzo (right).
    The highlighted regions indicate the corresponding microscopic views, revealing distinctive surface structure even on visually uniform materials.}
    \label{fig:macro_micro}
    \vspace{-8pt}
\end{figure}

High optical magnification provides a second benefit beyond increased geometric sensitivity: it can reveal natural surface structure that is not readily visible or distinctive at conventional imaging scales.
Surfaces that appear weakly textured at the macro scale can expose rich microstructure, while surfaces with visible macroscopic patterns can reveal additional fine-scale structure under microscopic observation.
Figure~\ref{fig:macro_micro} illustrates this change in appearance for the surfaces considered in this work.
We use this naturally occurring microtexture as a visual reference for localization without modifying the surface with artificial landmarks.

Using this sensing configuration, we show that natural microscopic surface texture supports globally referenced planar localization in the tens-of-micrometers regime across multiple surfaces.
At this scale, localization errors are a fraction of the diameter of a human hair.
This precision is achieved with a localization pipeline built on an established ground texture localization method, without requiring a localization algorithm specialized for micrometer-scale operation.

We envision this level of precision for localized high-precision work regions, where such accuracy is needed near the task rather than throughout the broader environment.
The system is therefore not intended to replace general-purpose navigation across an entire factory or building.

The primary contributions of this work are:
\begin{itemize}
    \item A geometric formulation relating optical magnification, image-space localization uncertainty, and physical localization precision, motivating the use of microscopic surface texture for high-precision localization
    
    \item A microscope-based ground texture localization system that addresses the calibration, camera-modeling, and focus challenges introduced by high-magnification optics
    
    \item A demonstration of tens-of-micrometers global localization on a low-cost mobile robot using a surface map collected independently with a precision gantry
\end{itemize}

We validate these contributions across three distinct surface types using a calibrated gantry reference system and further demonstrate our system's ability to measure micrometer-scale displacement by observing the thermal expansion of an aluminum beam.
We will release the resulting high-resolution microscopic ground texture datasets and accompanying code.

\section{Related Work}

\subsection{Conventional Robot Localization}

Modern mobile robots commonly use visual, LiDAR, visual-inertial, or multimodal localization and SLAM methods based on macroscopic environmental structure~\cite{campos_orb-slam3_2021, yuan_sdv-loam_2023}.
These approaches support robust localization across diverse environments, but their physical accuracy generally remains in the centimeter-to-millimeter regime, substantially coarser than the tens-of-micrometers precision considered here.
Applications requiring substantially finer positioning therefore motivate sensing and positioning approaches designed for a different precision regime~\cite{cornejo_industrial_2025, swyt_design_1995}.

\subsection{Surface Texture as a Visual Reference}

Ground texture localization uses the appearance of the surface beneath a robot as a visual reference.
Early work showed that floor appearance mosaics could support robot localization, establishing that the ground surface itself can serve as a map~\cite{kelly_mobile_2000}.
More recent work has pushed this idea toward higher precision, including the HD Ground benchmark, where localization methods demonstrated sub-millimeter accuracy using high-resolution ground imagery~\cite{schmid_hd_2022}.
L-GROUT demonstrated lightweight global metric localization on this benchmark using local visual features and geometric voting~\cite{wilhelm_lightweight_2024}, and GTBoW further showed that ground texture can support place recognition across long time intervals~\cite{wilhelm_improved_2025}.
These methods operate at fine spatial scales for robot localization, but use surface imagery at a coarser physical scale than the microscopic imagery considered here.

Related work on physical fingerprints further shows that natural materials contain distinctive visual structure at fine spatial scales.
Microscopic arrangements of fibers and other surface features have been used to identify nominally similar pieces of paper and other materials through physically unclonable visual signatures~\cite{gao_physical_2020, clarkson_fingerprinting_2009}.
Although this literature addresses identification rather than localization, it demonstrates that natural surface microstructure can provide distinctive visual information.

Digital image correlation (DIC) provides a related optical approach for high-precision displacement and strain measurement by correlating surface appearance between images~\cite{pan_two-dimensional_2009}.
Conventional 2D DIC estimates in-plane deformation of approximately planar surfaces and commonly uses applied speckle patterns, although natural surface texture can also provide sufficient image structure.
For example, small local strains have been measured directly from the natural texture of concrete~\cite{ferrer_method_2019}.
Unlike DIC, which typically measures deformation between corresponding images of the same region, our objective is to globally localize a query image within a prebuilt microscopic surface map.

\subsection{High-Precision Positioning Systems}

At finer precision scales, positioning is commonly achieved using dedicated mechanical or metrology systems.
CNC machines and precision gantries achieve micrometer-scale positioning through rigid mechanical structures and accurate proprioceptive feedback, but their precision is tied to a fixed work volume~\cite{chan_finite_2023}.
Coordinate Measuring Machines similarly use rigid frames and precision probing for highly accurate dimensional measurement~\cite{ali_probing_2010}.
These systems provide substantially greater precision than conventional robot localization, but depend on specialized hardware and mechanically constrained workspaces.

Artificial references provide another route to globally referenced positioning.
Fiducial markers and other calibrated visual landmarks simplify pose estimation by introducing detectable references at known locations and have been used for robot localization and docking~\cite{zhang_localization_2015}.
Such approaches can provide precise localization, but require placing and calibrating references in the environment.
In contrast, our approach uses natural surface appearance as the global reference and investigates whether microscopic imaging can extend ground texture localization to a finer precision regime.

\section{System Design and Methodology}

\subsection{Geometric Sensitivity and Microscopic Imaging}
\label{ref:geometric_sensitivity}

Ground texture localization is typically formulated using a downward-facing camera that observes a locally planar surface from an approximately fixed height.
With the imaging geometry calibrated, the image pose within the surface map can be represented in SE(2) by planar translation and rotation.

For a small physical displacement $\delta \mathbf{x}$ in the surface plane, the corresponding displacement $\delta \mathbf{u}$ in the native sensor image can be locally approximated as $\delta \mathbf{u} = \mathbf{J}\delta \mathbf{x} + \boldsymbol{\epsilon}$, where $\mathbf{J}$ is the local physical-to-image Jacobian and $\boldsymbol{\epsilon}$ represents image-space estimation error.
We refer to this relationship as the geometric sensitivity of the imaging system.
If the image-space estimate has covariance $\mathbf{\Sigma}_u$, the corresponding physical localization uncertainty is approximately
\begin{equation}
    \mathbf{\Sigma}_x \approx
    \mathbf{J}^{-1}\mathbf{\Sigma}_u\mathbf{J}^{-T}.
\end{equation}

This expression highlights two ways to improve physical localization precision.
First, increasing optical magnification increases the magnitude of $\mathbf{J}$, so a given physical displacement produces a larger displacement in the native sensor image.
Second, reducing $\mathbf{\Sigma}_u$ improves the precision with which that image displacement is estimated, with this uncertainty depending on the localization algorithm and image properties such as texture, focus, and blur.
We primarily act on the first of these factors by using microscope optics to increase optical magnification and therefore geometric sensitivity.

Under the approximately isotropic scaled-orthographic geometry used here, this geometric sensitivity can be expressed more directly as $\mathbf{J} \approx \rho \cdot \mathbf{I}$, where $\rho$ is the calibrated native-image scale in pixels per unit surface distance.
The corresponding one-dimensional relationship is $\sigma_x \approx \sigma_u/\rho$.
Thus, $\rho$ provides a direct measure of optical geometric sensitivity, while $\sigma_u$ captures how precisely the resulting image displacement can be estimated.
Importantly, $\rho$ is defined by the optical projection in native sensor coordinates rather than by the digital resolution at which an image is subsequently processed.
Digital downsampling can therefore affect localization through image-space estimation uncertainty and feature availability, but it does not change the underlying optical geometric sensitivity.

We also maintain consistent imaging conditions to limit variation in image-space estimation uncertainty.
On-board illumination provides consistent surface appearance, while exposure, analog gain, white-balance gains, and illumination are fixed for each physical camera so that these settings do not vary between images or surface textures.

\subsection{Camera Calibration and Autofocus}

High-magnification microscope optics have a shallow depth of field, so images are acquired at an approximately fixed distance from the locally planar surface.
Over this narrow operating range, the imaging geometry is well approximated by scaled-orthographic projection with approximately constant metric scale, reducing localization to planar translation and rotation in $\mathrm{SE}(2)$.

The shallow depth of field also makes conventional multi-view pinhole calibration impractical because strongly tilted calibration targets cannot remain uniformly in focus.
To retain compatibility with SLAM software that assumes pinhole intrinsics, we represent the scaled-orthographic calibration using a virtual pinhole model.
Under the locally planar surface model, all surface points are assigned a common virtual depth $Z_v$, and the virtual focal length $f_v$ is chosen such that $f_v/Z_v=\rho$, where $\rho$ is the calibrated native-image scale.
This reproduces the calibrated metric scale while allowing conventional pinhole-based software to operate on the microscopic imagery.
This virtual depth is introduced only to reproduce the calibrated scale and does not correspond to the physical camera-to-surface distance.

Each camera is calibrated using an external stage micrometer containing a grid with known physical spacing.
The calibration estimates the native metric image scale and lens distortion used to correct captured images.
These camera parameters are frozen before gantry calibration or localization experiments are performed, keeping the optical calibration independent from the subsequent mechanical calibration.

A challenge of microscopic imaging is the limited working-depth range, which makes autofocus necessary as the surface height varies.
We autofocus by translating the camera vertically through a coarse-to-fine search and evaluating image sharpness using the Brenner gradient metric.
For image sequences, an initial broad search establishes the local focal plane, while subsequent images use narrower searches centered near the previous valid focus estimate.
If a local search cannot confidently identify the focus peak, the search range is expanded and focus is reacquired.

Characterization of the gantry microscope established an approximately $\pm0.10$ mm usable range about the local best-focus plane.
We measured the calibrated image scale to change by approximately $-0.61\%$ per millimeter of camera-height change, so the accepted focus range corresponds to only about $\pm0.061\%$ variation in image scale.
This variation is sufficiently small that we use the nominal calibrated image scale rather than introducing a focus-dependent scale model for every image, with any remaining error from this approximation reflected in the empirical reference-system uncertainty described below.

\subsection{Calibrated Gantry Reference System}

\begin{figure}[t]
    \vspace{5pt}
    \centering
    \includegraphics[width=0.98\columnwidth]{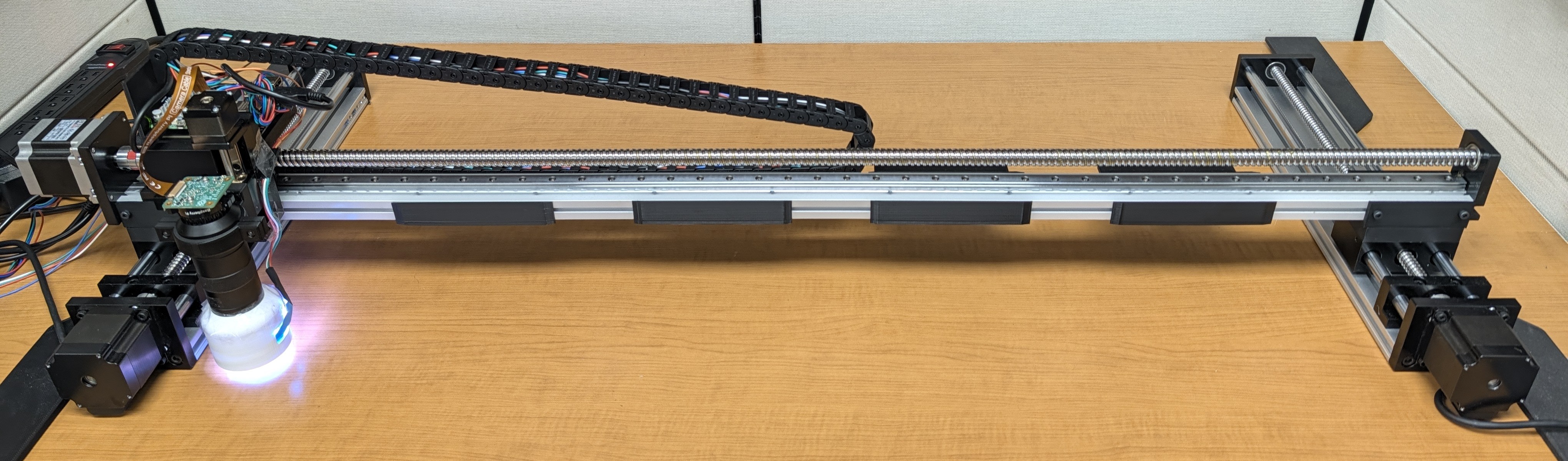}
    \caption{\textbf{Gantry-based Mapping and Reference System.}
    A microscope camera mounted to a calibrated XY gantry provides controlled motion for surface-map acquisition and independent localization queries.}
    \label{fig:gantry}
    \vspace{-8pt}
\end{figure}

We use a motorized XY gantry for map acquisition and reference measurement (Fig.~\ref{fig:gantry}).
Because this work operates in the tens-of-micrometers precision regime, the reference system must provide accuracy at a comparable or finer scale.
This makes many conventional robotics ground-truth systems unsuitable for the evaluation considered here, motivating the calibrated gantry-based approach used in this work.
Its controlled motion enables dense and repeatable collection of surface imagery with known image locations, while its calibrated geometry provides the reference coordinates used to evaluate localization accuracy.

Gantry geometry is calibrated using stainless-steel plates marked with a high-precision laser system and containing an Anoto-based coded-dot pattern~\cite{hostettler_real-time_2016}.
The coded pattern provides globally identifiable plate coordinates, allowing individual images to be localized directly relative to the plate.
The camera calibration described in Sec.~III-B is completed and frozen beforehand so that gantry calibration cannot alter the optical scale to absorb mechanical errors.
No stitched visual reference map is required, avoiding additional image-registration errors.

The final gantry model estimates an isotropic motion scale, gantry-axis nonorthogonality, and the planar yaw between the camera image frame and gantry coordinate frame.
We also evaluated independent scale factors for the two gantry axes, but the calibration data did not justify the additional parameters, so the simpler isotropic model was retained.
The camera is mounted approximately normal to the surface, allowing the gantry calibration to remain a planar geometric model.
Because these parameters depend on the physical gantry and camera mounting, the gantry is recalibrated whenever the system is relocated.
The resulting gantry parameters are frozen before dataset collection.

At tens-of-micrometers precision, mechanical and environmental effects that are negligible at coarser scales become measurable.
We therefore account for potential effects from backlash, settling, drift, temperature, warm-up, and thermal transients.
Backlash is mitigated through consistent approach directions, images are captured only after settling, and motor and illumination conditions are kept consistent during calibration and data collection.
Rather than introducing poorly constrained corrections for the remaining effects, we evaluate the frozen calibration on separate held-out measurements so that residual mechanical, thermal, and optical errors are reflected in the empirical reference uncertainty reported in Sec.~IV.



\subsection{Map Acquisition}

Surface maps are collected by moving the gantry through a regular overlapping grid.
At each location, the gantry stops, autofocus is performed, and a full-resolution image is captured using the fixed camera settings.
Each image is assigned a pose directly from the calibrated gantry reference model rather than from image stitching, visual SLAM, or feature-based map reconstruction.
This prevents visual-registration error from entering the reference map.

\subsection{Ground Texture Database and Localization}

Global localization builds on L-GROUT~\cite{wilhelm_lightweight_2024}, which we summarize briefly for completeness.
SIFT features~\cite{lowe_distinctive_2004} are extracted from each posed database image, after which L-GROUT retains a subset of 50 features per image and compresses their descriptors using principal component analysis (PCA) before storing them in a size-binned approximate nearest neighbor (ANN) database.

L-GROUT bins features according to their SIFT keypoint size, which both restricts ANN retrieval and provides a geometric scale filter for candidate matches.
This works naturally when map and query images are collected using the same camera configuration because their image-space feature sizes are directly comparable.
In this work, however, the gantry and mobile platforms operate at different native image scales.
We therefore convert SIFT keypoint diameters to physical units using each camera's calibrated image scale.
Because corresponding SIFT features can be assigned different characteristic scales, retrieved candidates are retained only when their physical scales satisfy $\left|\log(s_q/s_m)\right|\leq0.15$, where $s_q$ and $s_m$ are the physical query and map feature scales, respectively.
The nearest scale-compatible descriptor match then contributes a single pose vote.

The resulting votes are filtered by identifying the largest spatially and orientationally consistent cluster using a KD-tree.
RANSAC is then applied to the corresponding feature matches to estimate the final global $\mathrm{SE}(2)$ query pose.

\subsection{Mobile Platform}

To demonstrate localization on an independent mobile platform, we use a custom differential-drive robot equipped with a downward-facing camera and microscope lens, controlled illumination, a motorized focus stage, onboard computing, and motor-control hardware.
The robot is constructed from low-cost components, with a total cost below \$600.

The robot is used as a query platform against maps collected by the gantry.
To increase focus tolerance, the mobile microscope is operated at lower optical magnification than the gantry microscope.
The two platforms therefore have different geometric sensitivities and native image scales, and each uses its own independently calibrated camera model.

For the experiments in this work, the robot uses a stationary-capture acquisition procedure in which it moves to a query location, stops, autofocuses, and captures an image.
This was chosen to maximize image quality for the precision experiments.
Continuous-motion localization is also possible, but introduces an additional tradeoff among robot speed, exposure time, optical magnification, and motion blur that is discussed later in Sec.~\ref{sec:discussion}.

\section{Experiments and Results}

\subsection{Experimental Setup}

We evaluate the proposed system on three surface textures: a printed wood-grain desk surface, terrazzo, and concrete.
For each surface, we independently calibrate the gantry reference system and collect a localization database followed by a separate query set.
The resulting mapped region spans approximately $83.5\times30.4~\mathrm{cm}$, corresponding to roughly 3,700 non-overlapping native camera footprints.

Each surface is evaluated using the same set of 1,007 query poses spatially distributed throughout the mapped region.
Database and query acquisition are performed within the same calibrated gantry reference frame for each surface.

All gantry images are acquired at $4056\times3040$ pixels with a calibrated native image scale of approximately 424~pixels/mm.
The gantry reference geometry is calibrated independently for each surface.

Unless otherwise stated, localization extracts up to 1,000 SIFT features per query image, while L-GROUT retains 50 features per database image.
SIFT descriptors are reduced from 128 to 16 dimensions using PCA before approximate nearest-neighbor retrieval.
The physical feature-scale compatibility test described in Sec.~III uses a logarithmic scale tolerance of $\tau=0.15$.
Pose votes are clustered using a 0.4~mm spatial radius and a $10^\circ$ orientation tolerance, and the final pose is estimated with RANSAC using a $200~\mu$m inlier threshold and a minimum of three inliers.

In several experiments, we additionally use direct photometric registration using Enhanced Correlation Coefficient (ECC) optimization as a diagnostic reference to help distinguish L-GROUT-specific localization error from discrepancies also supported by the underlying imagery.
The photometric estimate is computed independently of L-GROUT, providing a separate measure of image alignment.
Because both methods operate on the same visual data, we do not treat the photometric estimate as independent ground truth.
Instead, agreement between the two suggests that discrepancies relative to the gantry reference are not specific to L-GROUT alone.
The experiment-specific initialization and search procedure are described separately where this reference is used.

We report localization success rate, planar translation error, and orientation error.
A query is considered successfully localized when L-GROUT returns a pose estimate.
Translation and orientation errors are computed relative to the reference pose as the Euclidean planar distance and wrapped heading difference, respectively.
Because the gantry provides a calibrated reference rather than exact ground truth, these errors are interpreted alongside its held-out calibration residuals reported in Sec.~\ref{sec:gantry_validation}.

\subsection{Gantry Reference Validation}
\label{sec:gantry_validation}

We quantify the accuracy of the calibrated gantry reference for each texture using held-out validation, where one calibration plate is excluded at a time.
Table~\ref{tab:gantry_calibration} reports the translation residuals aggregated across all held-out measurements for each texture.

\begin{table}[t]
\vspace*{5pt}
\centering
\caption{Held-out validation of the per-texture gantry reference-frame calibration.}
\label{tab:gantry_calibration}
\begin{tabular}{l cc}
\toprule
Texture & \shortstack{Held-out\\RMSE ($\mu$m)} & \shortstack{Held-out\\P95 ($\mu$m)} \\
\midrule
Printed Desk & 25.71 & 54.03 \\
Terrazzo & 30.97 & 68.22 \\
Concrete & 26.10 & 55.51 \\
\bottomrule
\end{tabular}
\vspace{-8pt}
\end{table}

The held-out gantry calibration achieves RMSEs of 25.71--30.97~$\mu$m and 95th-percentile errors of 54.03--68.22~$\mu$m across the three surfaces.
These residuals characterize the uncertainty of the gantry reference at a scale comparable to the localization errors reported below.
As a result, some portion of the measured localization error may arise from the reference system itself, rather than from L-GROUT alone.
We therefore treat the gantry as a calibrated reference rather than exact ground truth.

\subsection{Global Localization Across Surface Types}

\begin{table*}[t]
\vspace*{5pt}
\centering
\caption{Global localization across three surfaces with 1,007 gantry queries per surface. L-GROUT errors are reported relative to the calibrated gantry reference. The photometric reference provides an additional image-based comparison evaluated on queries for which it returns a valid pose, with \(N\) denoting the number of such queries.}
\label{tab:main_localization}
\resizebox{\textwidth}{!}{%
\begin{tabular}{l c | ccc | ccc}
\toprule
& &
\multicolumn{3}{c|}{L-GROUT vs. Gantry} &
\multicolumn{3}{c}{Photometric Reference Comparison} \\
Texture
& \shortstack{L-GROUT\\Success (\%)}
& RMSE ($\mu$m)
& Median ($\mu$m)
& \shortstack{Orient.\\RMSE ($^\circ$)}
& \shortstack{Photo. Ref.\\$N$}
& \shortstack{L-GROUT--Photo. Ref.\\RMSE ($\mu$m)}
& \shortstack{Photo. Ref.--Gantry\\RMSE ($\mu$m)} \\
\midrule
Printed Desk & 100.0 & 28.36 & 19.78 & 0.070 & 988 & 8.82 & 28.29 \\
Terrazzo & 100.0 & 57.08 & 50.16 & 0.091 & 997 & 8.86 & 57.69 \\
Concrete & 100.0 & 72.75 & 61.07 & 0.093 & 998 & 8.79 & 72.86 \\
\bottomrule
\end{tabular}%
}
\vspace{-8pt}
\end{table*}

We next evaluate global localization across the printed desk, terrazzo, and concrete surfaces using the held-out gantry queries described above.
Each query is localized against the corresponding surface database; the gantry pose is not provided to L-GROUT and is used only for evaluation.
Table~\ref{tab:main_localization} reports localization success and pose error relative to the calibrated gantry reference.

L-GROUT returns a pose for every query on all three surfaces, corresponding to a 100\% localization success rate.
Translation RMSE ranges from 28.36--72.75~$\mu$m across the three surfaces, while orientation RMSE remains below $0.1^\circ$ in all cases.
For the printed desk, the translation RMSE is on the same scale as the held-out gantry reference uncertainty, while the terrazzo and concrete errors are larger than their corresponding reference uncertainties.
We therefore use the photometric reference below to examine whether these larger discrepancies are specific to L-GROUT or are also supported by direct image alignment.

To provide an additional image-based comparison, we compute a local photometric reference using direct ECC registration.
For each query, the gantry reference is used only to select nearby overlapping database images and provide the starting alignment for ECC registration.
After rejecting low-quality or spatially inconsistent registrations, the retained estimates are robustly aggregated into a single photometric reference pose.
The photometric reference does not return a valid pose for every query, so comparisons involving it are evaluated only on the subset for which a reference pose is available; the corresponding sample count is reported as \(N\) in Table~\ref{tab:main_localization}.

L-GROUT and the photometric reference agree within 8.79--8.86~$\mu$m RMSE across all three surfaces.
Moreover, the photometric-reference--gantry errors closely match the corresponding L-GROUT--gantry errors.
Taken together, these results show that L-GROUT closely follows the pose supported by direct image alignment, while the larger gantry-referenced discrepancies on terrazzo and concrete are largely shared by both visual estimates rather than being specific to L-GROUT.

\subsection{Orientation Robustness}
\label{sec:orientation_robustness}

\begin{figure}[t]
    \centering
    \includegraphics[width=0.98\columnwidth]{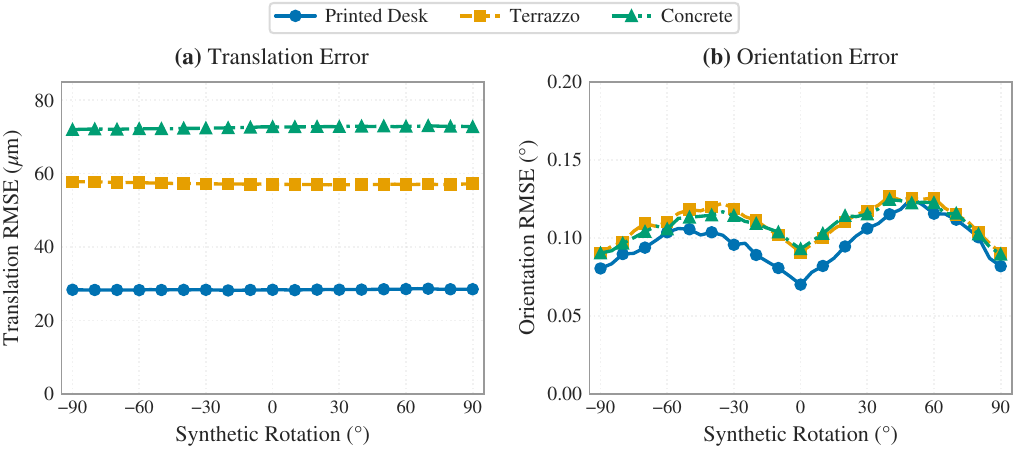}
    \caption{\textbf{Robustness to Query-image Rotation.}
    Gantry query images are synthetically rotated from $-90^\circ$ to $+90^\circ$.
(a) Translation RMSE and (b) orientation RMSE remain stable across the tested range.}
    \label{fig:rotation_robustness}
\end{figure}

The gantry provides precise translational motion but does not support controlled camera rotation.
To verify that localization performance is not dependent on near-aligned query and map headings, we synthetically rotate each held-out query from \(-90^\circ\) to \(90^\circ\) in \(5^\circ\) increments.
Each image is rotated about the calibrated principal point, cropped to valid image content, and evaluated using a correspondingly adjusted reference heading.

L-GROUT localizes all 1,007 queries at every tested rotation for all three surfaces.
Translation RMSE remains nearly constant across the full range, varying from 28.19--28.67~$\mu$m for the printed desk, 56.95--57.80~$\mu$m for terrazzo, and 72.00--73.04~$\mu$m for concrete.
Orientation RMSE remains below approximately $0.127^\circ$ for every surface and rotation.
These results show that the localization performance is not dependent on near-aligned map and query imagery, and that the reported precision is maintained across large heading differences.

\subsection{Image Resolution and Runtime}

\begin{figure}[t]
    \centering
    \includegraphics[width=0.98\columnwidth]{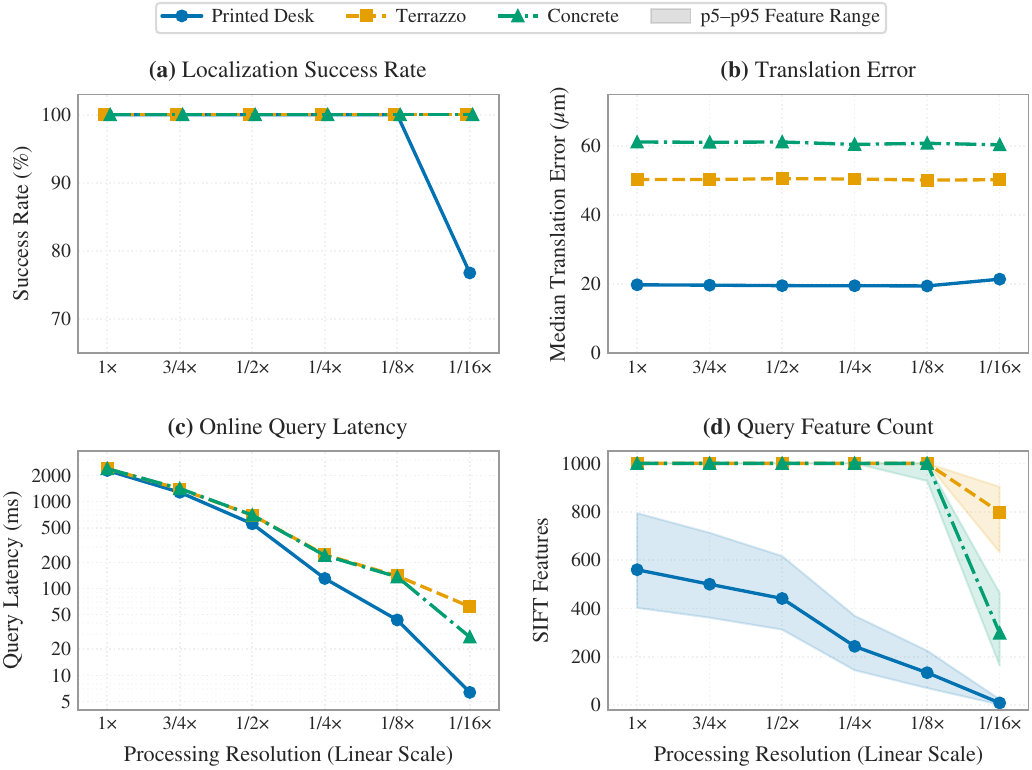}
    \caption{\textbf{Image Resolution Tradeoffs.}
Effect of digital downsampling on (a) localization success, (b) median translation error, (c) single-worker query latency, and (d) retained SIFT features.
Precision remains stable until texture-dependent feature loss reduces localization success.
  }
    \label{fig:resolution_tradeoff}
    \end{figure}

We next examine how much the digital image resolution can be reduced while preserving the localization precision enabled by the microscope optics.
To evaluate the tradeoff between digital resolution, localization performance, and computational cost, the same database and query images are symmetrically downsampled from their native \(4056\times3040\) resolution to \(1/16\times\) linear resolution while preserving the physical field of view.

As shown in Fig.~\ref{fig:resolution_tradeoff}(a) and (b), localization success remains 100\% for all three surfaces through $1/8\times$ resolution, while median translation error remains nearly unchanged.
At $1/8\times$, median errors are 19.44~$\mu$m for the printed desk, 50.13~$\mu$m for terrazzo, and 60.81~$\mu$m for concrete, compared with 19.77, 50.30, and 61.17~$\mu$m at native resolution.
At $1/16\times$, terrazzo and concrete remain fully successful, while printed-desk success falls to 76.8\%.
Among successful printed-desk queries, the median error remains similar at 21.41~\(\mu\)m.

Downsampling also substantially reduces computational cost.
Median online query latency decreases from approximately 2.3--2.4~s at native resolution ($\sim$0.4~Hz) to 43.45~ms for the printed desk, 139.53~ms for terrazzo, and 136.12~ms for concrete at $1/8\times$, corresponding to approximately 23, 7.2, and 7.3~Hz, respectively.
Thus, substantial computational savings are possible before localization performance begins to degrade.

Figure~\ref{fig:resolution_tradeoff}(d) helps explain the printed-desk failures at the most aggressive downsampling.
Its median SIFT feature count decreases from 560 at native resolution to 134 at $1/8\times$ and only 9 at $1/16\times$, whereas terrazzo and concrete still retain medians of 798 and 298 features, respectively, at $1/16\times$.
The drop in printed-desk success at $1/16\times$ is primarily associated with too few feature correspondences to form a valid pose estimate, while the accuracy of successfully returned estimates remains largely unchanged.

These results are consistent with the uncertainty relationship in Sec.~III-A, since digital downsampling does not change the underlying optical projection that determines geometric sensitivity.
Across the tested range, localization precision remains nearly unchanged despite substantial reductions in digital image resolution.
This suggests that the optical geometric sensitivity is preserved under downsampling, allowing large computational savings until too much visual information is lost to maintain reliable localization.

\subsection{Cross-Platform Mobile Localization}

We next evaluate whether maps acquired with the gantry can be used for localization from a separate mobile platform.
For each surface, the robot follows trajectories containing a total of 306 query images.
Unlike the synthetic orientation experiment in Sec.~IV-D, these trajectories physically vary the robot heading while also introducing a separately calibrated camera and different microscope magnification.

For evaluation, we generate a map-relative photometric reference for each trajectory using direct ECC registration against the gantry database images.
The robot's open-loop commanded trajectory provides a coarse seed for the local image search, and a reference pose is retained only when the registrations satisfy overlap, correlation, and consistency checks.
Because this reference is generated independently of L-GROUT but relies on the same visual data, we use it as an image-based comparison rather than independent ground truth.

\begin{table}[t]
\vspace*{5pt}
\centering
\caption{Mobile localization evaluated using the map-relative photometric reference.
Each surface contains 306 queries across three trajectories.
Errors are computed on the shared subset for which both L-GROUT and the photometric reference return a pose, with Evaluation $N$ denoting the size of this subset.}
\label{tab:mobile_localization}

\resizebox{\columnwidth}{!}{%
\begin{tabular}{l cc | ccc}
\toprule
& \multicolumn{2}{c|}{Evaluation Coverage}
& \multicolumn{3}{c}{L-GROUT vs. Photometric Reference} \\
Texture
& \shortstack{Photometric Reference\\Coverage (\%)}
& \shortstack{L-GROUT\\Success (\%)}
& \shortstack{Evaluation\\$N$}
& \shortstack{Trans. RMSE\\($\mu$m)}
& \shortstack{Orient. RMSE\\($^\circ$)} \\
\midrule
Printed Desk & 88.2 & 100.0 & 270 & 20.70 & 0.067 \\
Terrazzo & 99.3 & 100.0 & 304 & 26.42 & 0.101 \\
Concrete & 82.7 & 98.4 & 252 & 22.67 & 0.135 \\
\bottomrule
\end{tabular}%
}
\vspace{-8pt}
\end{table}

As shown in Table~\ref{tab:mobile_localization}, L-GROUT localizes all queries on the printed desk and terrazzo and 98.4\% of the concrete queries.
The photometric reference provides valid poses for 82.7--99.3\% of the queries across the three surfaces, defining the subsets used for quantitative comparison.
Over these subsets, translation RMSE relative to the photometric reference remains between 20.70 and 26.42~$\mu$m, with orientation RMSE below $0.14^\circ$ for all three surfaces.

These results show that gantry-built microscopic surface maps can be reused by a separately calibrated mobile platform despite differences in camera calibration, microscope magnification, and physical heading.
The close agreement with the photometric reference further indicates that this cross-platform transfer preserves consistent localization relative to the gantry-built map.

\subsection{Micrometer-Scale Thermal Expansion Measurement}

    \begin{figure}[t]
    \centering
    \vspace{5pt}
    \begin{subfigure}[b]{0.98\columnwidth}
        \centering
        \includegraphics[width=\linewidth]{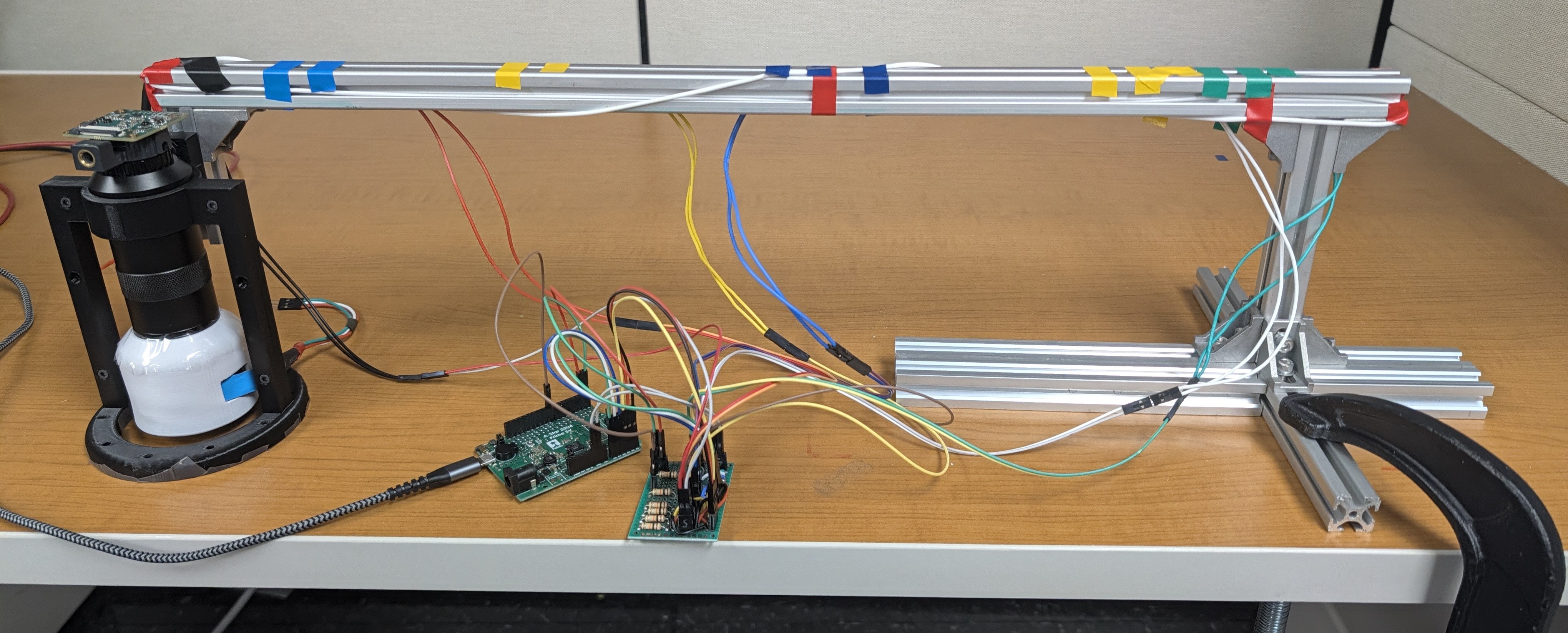}
    \end{subfigure}

    \vspace{2pt}

    \begin{subfigure}[b]{0.49\columnwidth}
        \centering
        \includegraphics[width=\linewidth]{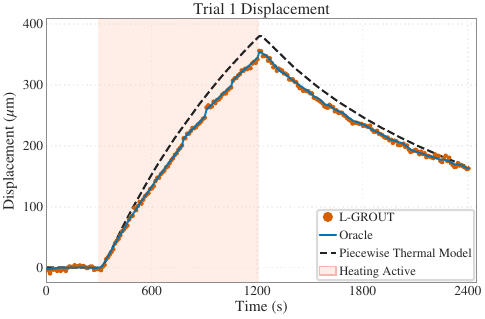}
    \end{subfigure}
    \hfill
    \begin{subfigure}[b]{0.49\columnwidth}
        \centering
        \includegraphics[width=\linewidth]{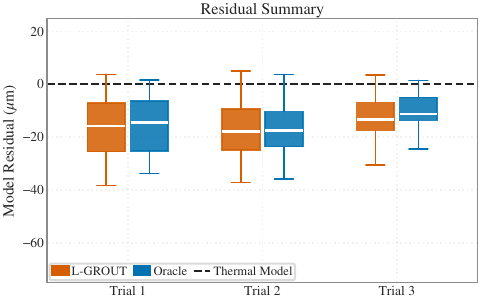}
    \end{subfigure}

\caption{\textbf{Thermal Expansion Measurement.}
Top: Experimental setup used to measure thermal expansion of the aluminum rod.
Bottom-left: Representative displacement measurements from Trial 1, showing L-GROUT and direct ECC estimates compared with the thermal model, with shading indicating the heating interval.
Bottom-right: Residual distributions relative to the thermal model across all three trials.}
    \label{fig:thermal_expansion}
    \vspace{-8pt}
\end{figure}

Finally, we perform a thermal-expansion sensing experiment to demonstrate the system's ability to measure micrometer-scale displacement.
In the setup shown in Fig.~\ref{fig:thermal_expansion}, we clamp one end of a 53.6~cm free span of an aluminum rod and observe the textured surface near its free end using a fixed microscope camera.
A heating wire is wrapped around the rod, which is thermally insulated to reduce heat loss and promote a more uniform temperature profile.
Six thermistors distributed along the rod measure the remaining spatial temperature variation.
The expected axial expansion is estimated from a piecewise-linear temperature profile according to
\begin{equation}
    \Delta L(t)
    =
    \alpha \int_0^L \Delta T(x,t)\,dx,
\end{equation}
where $L$ is the free rod length and $\alpha = 23.1\times10^{-6}~^\circ\mathrm{C}^{-1}$ is the coefficient of thermal expansion used for aluminum.

We perform three trials consisting of baseline, heating, and cooling periods, with images captured every 10~s.
For each image, L-GROUT estimates displacement using the existing gantry-built map.
We additionally use direct ECC registration to a fixed cold reference image to obtain a separate image-based displacement estimate.

The representative trial shown in the bottom-left panel of Fig.~\ref{fig:thermal_expansion} illustrates that both L-GROUT and the direct ECC displacement estimate closely follow the temperature-based expansion model.
Across the three trials, ECC measures a mean total axial excursion of $358.8\pm5.4~\mu$m, and linear fits to the thermal prediction yield $R^2>0.996$ in every trial.
L-GROUT returns a pose for every image and agrees with the ECC estimate within 2.57--3.64~$\mu$m RMSE.
The residual distributions across all three trials are summarized in the bottom-right panel.

Together, these trials demonstrate repeatable measurement of approximately 360~$\mu$m of thermal expansion while resolving micrometer-scale displacement throughout the heating and cooling cycle.

\section{Discussion}
\label{sec:discussion}

\textbf{Reference Precision.}
Evaluating localization at the tens-of-micrometers scale places unusually stringent requirements on the reference system itself.
Held-out validation of the calibrated gantry gives translation RMSEs of approximately $26$--$31~\mu$m, comparable to some of the localization errors reported here.
The close agreement between L-GROUT and the separate photometric estimates further suggests that part of the gantry-referenced discrepancy is not specific to the localization algorithm.
We therefore interpret the reported localization errors in the context of the measured gantry uncertainty rather than treating the gantry as exact ground truth.
At this scale, mechanical, optical, thermal, and other environmental effects that are negligible in conventional robot-localization experiments can become measurable, making further separation of localization error from reference-system error increasingly difficult.

\textbf{Map Scale.}
The mapped region covers approximately $83.5\times30.4~\mathrm{cm}$, corresponding to $\sim$3,700 non-overlapping camera footprints of approximately $9.57\times7.17~\mathrm{mm}$.
For comparison, the largest HD Ground map covers $106.12~\mathrm{m}^2$ with an approximately $16.7\times12.5~\mathrm{cm}$ footprint, or roughly 5,100 nominal footprints.
Thus, despite its smaller physical area, our map is of the same order in extent when measured in camera footprints.

L-GROUT also provides a compact map representation, retaining 50 PCA-16 features per database image, or 3.2~kB of descriptor data using 32-bit floats.
Including feature poses, physical scales, size-bin indices, and database metadata, each 12,936-image surface map occupies approximately 86~MiB, compared with approximately 27--35~GiB for the full-resolution source imagery.
This leaves substantial storage headroom for larger microscopic maps, although the effects of further database growth on retrieval accuracy and latency remain to be characterized.

\textbf{Operating Constraints.}
The increased geometric sensitivity provided by microscope optics introduces several practical constraints.
High magnification produces a shallow depth of field, making the system sensitive to surface-height variations and motivating the autofocus procedure.
Motion during exposure can also introduce blur, so the mobile experiments use stationary image capture to ensure high-quality query images.
Stationary acquisition is not required by the localization method, but continuous-motion operation would require balancing robot speed, exposure time, illumination, optical magnification, and focus tolerance.
Stronger illumination and shorter exposures can reduce motion blur, while lower magnification increases depth of field and motion tolerance at the cost of geometric sensitivity.

\textbf{Relation to Deformation Measurement.}
The ability to resolve micrometer-scale displacement also motivates combining global microscopic localization with digital image correlation, which estimates local displacement and strain from changes in surface appearance.
Global localization could return a mobile platform to mapped material regions, where local image correlation could measure deformation or strain for in-situ monitoring using natural surface texture.

\section{Conclusion}

This work demonstrates that natural microscopic surface texture can provide a global reference for planar localization in the tens-of-micrometers regime.
By increasing optical magnification, ground texture localization can operate at a substantially finer physical scale while retaining successful global matching across multiple surface types.
Using a calibrated gantry reference, we demonstrate high-precision global localization across multiple surfaces.
We further show cross-platform map reuse on a mobile robot and repeatable micrometer-scale displacement measurement through a thermal-expansion experiment.
More broadly, these results indicate that microscopic surface texture can bring mobile localization into a precision regime typically associated with fixed positioning systems.





\section*{Acknowledgment}

Generative AI (ChatGPT and Claude) was used to assist with writing code and the manuscript. All outputs were reviewed by the authors.


\bibliographystyle{IEEEtran}
\bibliography{IEEEcontrol,references}

\begin{thebibliography}{10}
\providecommand{\url}[1]{#1}
\csname url@samestyle\endcsname
\providecommand{\newblock}{\relax}
\providecommand{\bibinfo}[2]{#2}
\providecommand{\BIBentrySTDinterwordspacing}{\spaceskip=0pt\relax}
\providecommand{\BIBentryALTinterwordstretchfactor}{4}
\providecommand{\BIBentryALTinterwordspacing}{\spaceskip=\fontdimen2\font plus
\BIBentryALTinterwordstretchfactor\fontdimen3\font minus \fontdimen4\font\relax}
\providecommand{\BIBforeignlanguage}[2]{{%
\expandafter\ifx\csname l@#1\endcsname\relax
\typeout{** WARNING: IEEEtran.bst: No hyphenation pattern has been}%
\typeout{** loaded for the language `#1'. Using the pattern for}%
\typeout{** the default language instead.}%
\else
\language=\csname l@#1\endcsname
\fi
#2}}
\providecommand{\BIBdecl}{\relax}
\BIBdecl

\bibitem{campos_orb-slam3_2021}
C.~Campos, R.~Elvira, J.~J. Rodriguez, J.~M. Montiel, and J.~D. Tardos, ``{ORB}-{SLAM3}: {An} {Accurate} {Open}-{Source} {Library} for {Visual}, {Visual}-{Inertial}, and {Multimap} {SLAM},'' \emph{IEEE Transactions on Robotics}, vol.~37, no.~6, pp. 1874--1890, Dec. 2021, arXiv: 2007.11898.

\bibitem{yuan_sdv-loam_2023}
Z.~Yuan, Q.~Wang, K.~Cheng, T.~Hao, and X.~Yang, ``{SDV}-{LOAM}: {Semi}-{Direct} {Visual}–{LiDAR} {Odometry} and {Mapping},'' \emph{IEEE Transactions on Pattern Analysis and Machine Intelligence}, vol.~45, no.~9, pp. 11\,203--11\,220, Sep. 2023.

\bibitem{cornejo_industrial_2025}
J.~Cornejo, ``Industrial, {Collaborative}, and {Mobile} {Robotics} in {Aerospace} {Manufacturing}: {Technical} {Review} focused on {Industry} 5.0,'' in \emph{2025 9th {International} {Symposium} on {Multidisciplinary} {Studies} and {Innovative} {Technologies} ({ISMSIT})}, Nov. 2025, pp. 1--6.

\bibitem{swyt_design_1995}
D.~A. Swyt, ``\BIBforeignlanguage{en}{Design, specification and tolerancing of micrometer-tolerance assemblies},'' National Institute of Standards and Technology, Gaithersburg, MD, Tech. Rep. NIST IR 5615, 1995.

\bibitem{kelly_mobile_2000}
A.~Kelly, ``Mobile {Robot} {Localization} from {Large}-{Scale} {Appearance} {Mosaics},'' Tech. Rep., 2000.

\bibitem{schmid_hd_2022}
J.~F. Schmid, S.~F. Simon, R.~Radhakrishnan, S.~Frintrop, and R.~Mester, ``{HD} {Ground} - {A} {Database} for {Ground} {Texture} {Based} {Localization},'' in \emph{2022 {International} {Conference} on {Robotics} and {Automation} ({ICRA})}.\hskip 1em plus 0.5em minus 0.4em\relax IEEE, May 2022, pp. 7628--7634.

\bibitem{wilhelm_lightweight_2024}
A.~Wilhelm and N.~Napp, ``Lightweight {Ground} {Texture} {Localization},'' in \emph{2024 {IEEE} {International} {Conference} on {Robotics} and {Automation} ({ICRA})}.\hskip 1em plus 0.5em minus 0.4em\relax IEEE, May 2024, pp. 10\,223--10\,229.

\bibitem{wilhelm_improved_2025}
------, ``Improved {Bag}-of-{Words} {Image} {Retrieval} with {Geometric} {Constraints} for {Ground} {Texture} {Localization},'' in \emph{2025 {IEEE} {International} {Conference} on {Robotics} and {Automation} ({ICRA})}, May 2025, pp. 8020--8026.

\bibitem{gao_physical_2020}
Y.~Gao, S.~F. Al-Sarawi, and D.~Abbott, ``\BIBforeignlanguage{en}{Physical unclonable functions},'' \emph{\BIBforeignlanguage{en}{Nature Electronics}}, vol.~3, no.~2, pp. 81--91, Feb. 2020.

\bibitem{clarkson_fingerprinting_2009}
W.~Clarkson, T.~Weyrich, A.~Finkelstein, N.~Heninger, J.~A. Halderman, and E.~W. Felten, ``Fingerprinting blank paper using commodity scanners,'' in \emph{Proceedings - {IEEE} {Symposium} on {Security} and {Privacy}}, 2009, pp. 301--314.

\bibitem{pan_two-dimensional_2009}
B.~Pan, K.~Qian, H.~Xie, and A.~Asundi, ``Two-dimensional digital image correlation for in-plane displacement and strain measurement: a review,'' \emph{Measurement Science and Technology}, vol.~20, no.~6, p. 062001, Jun. 2009.

\bibitem{ferrer_method_2019}
B.~Ferrer, J.~Espinosa, and D.~Mas, ``\BIBforeignlanguage{en}{A method to measure small local strains in concrete surfaces using its natural texture and image cross‐correlation},'' \emph{\BIBforeignlanguage{en}{Structural Control and Health Monitoring}}, vol.~26, no.~9, Sep. 2019.

\bibitem{chan_finite_2023}
T.-C. Chan, A.~Ullah, B.~Roy, and S.-L. Chang, ``Finite element analysis and structure optimization of a gantry-type high-precision machine tool,'' \emph{Scientific Reports}, vol.~13, p. 13006, Aug. 2023.

\bibitem{ali_probing_2010}
S.~Ali, ``\BIBforeignlanguage{en}{Probing {System} {Characteristics} in {Coordinate} {Metrology}},'' \emph{\BIBforeignlanguage{en}{Measurement Science Review}}, vol.~10, no.~4, Jan. 2010.

\bibitem{zhang_localization_2015}
H.~Zhang, C.~Zhang, W.~Yang, and C.~Chen, \emph{Localization and navigation using {QR} code for mobile robot in indoor environment}, Dec. 2015, pages: 2506.

\bibitem{hostettler_real-time_2016}
L.~Hostettler, A.~Ozgur, S.~Lemaignan, P.~Dillenbourg, and F.~Mondada, ``Real-time high-accuracy {2D} localization with structured patterns,'' in \emph{Proceedings - {IEEE} {International} {Conference} on {Robotics} and {Automation}}, vol. 2016-June.\hskip 1em plus 0.5em minus 0.4em\relax Institute of Electrical and Electronics Engineers Inc., Jun. 2016, pp. 4536--4543.

\bibitem{lowe_distinctive_2004}
D.~G. Lowe, ``Distinctive {Image} {Features} from {Scale}-{Invariant} {Keypoints},'' Tech. Rep., 2004, publication Title: International Journal of Computer Vision Volume: 60 Issue: 2.

\end{thebibliography}

\end{document}